\documentclass[10pt,twocolumn,letterpaper]{article}

\usepackage{wacv}
\usepackage{times}
\usepackage{epsfig}
\usepackage{graphicx}
\usepackage{amsmath}
\usepackage{amssymb}
\usepackage{booktabs}
\usepackage{enumerate}
\usepackage{enumitem}
\usepackage{soul}
\usepackage{booktabs}
\usepackage{multirow}
\usepackage{algorithm}
\usepackage{algorithmic}
\usepackage{makecell}
\usepackage{subcaption}

\wacvfinalcopy
\def\wacvPaperID{1501} % Enter the WACV Paper ID here

\wacvalgorithmstrack   % Uncomment this line if you are submitting to the Algorithms Track.
\ifwacvfinal
\usepackage[breaklinks=true,bookmarks=false]{hyperref}
\else
\usepackage[pagebackref=true,breaklinks=true,colorlinks,bookmarks=false]{hyperref}
\fi

\begin{document}

%%%%%%%%% TITLE
\title{SkeletonMAE: Spatial-Temporal Masked Autoencoders for Self-supervised Skeleton Action Recognition}

% \and

% \author{
% Wenhan Wu\\
% University of North Carolina at Charlotte\\
% {\tt\small wwu25@uncc.edu}
% \and
% Yilei Hua\\
% Wuhan University of Science and Technology\\
% % First line of institution2 address\\
% {\tt\small hyl9797510@gmail.com}
% \and 
% Ce Zheng\\
% University of Central Florida\\
% {\tt\small cezheng@knights.ucf.edu}
% \and
% Shiqian Wu\\
% Wuhan University of Science and Technology\\
% {\tt\small shiqian.wu@wust.edu.cn}
% \and
% Chen Chen\\
% University of Central Florida\\
% % First line of institution2 address\\
% {\tt\small chen.chen@crcv.ucf.edu}
% \and 
% Aidong Lu\\
% University of North Carolina at Charlotte\\
% {\tt\small wwu25@uncc.edu}
% }

\author{Wenhan Wu$^{1}$, Yilei Hua$^{2}$, Ce Zheng$^{3}$, Shiqian Wu$^{2}$, Chen Chen$^{3}$,  Aidong Lu$^{1}$\\
$^1$Department of Computer Science, University of North Carolina at Charlotte, USA\\
$^2$School of Information Science and Engineering, Wuhan University of Science and Technology, China\\
$^3$Center for Research in Computer Vision, University of Central Florida, USA\\
{\tt\small \{wwu25,alu1\}@uncc.edu;}
{\tt\small hyl9797510@gmail.com}\\
{\tt\small cezheng@knights.ucf.edu; shiqian.wu@wust.edu.cn; chen.chen@crcv.ucf.edu}
}

\maketitle
\thispagestyle{empty}

%%%%%%%%% ABSTRACT
\begin{abstract}
% Recently, self-supervised learning attracts more attention in skeleton-based action recognition and achieves striking results. 

Fully supervised skeleton-based action recognition has achieved great progress with the blooming of deep learning techniques. However, these methods require sufficient labeled data which is not easy to obtain. In contrast, self-supervised skeleton-based action recognition has attracted more attention. With utilizing the unlabeled data, more generalizable features can be learned to alleviate the overfitting problem and reduce the demand of massive labeled training data. 
% Inspired by the MAE \cite{he2022masked}, we propose a Spatial-Temporal Masked Autoencoder framework for self-supervised 3D skeleton-based action recognition (SkeletonMAE). Following a specific masking strategy, our pre-training method randomly mask frames and joints from the skeleton data and reconstruct the masked sequences.
% First, we introduce a novel masking strategy named Spatial-Temporal Masking for the skeleton data, which is applied in terms of both joint-level and frame-level. 
% % At the same time, we find that different joint-masking ratio and frame-masking ratio play significant roles in the SkeletonMAE.
% Second, we present a skeleton-based encoder-decoder architecture. An encoder is loaded to only process the unmasked sequences and a decoder is applied to model the reconstructed skeleton sequences during pre-training, which yields a comprehensive reconstruction of skeleton data to boost the self-supervised skeleton action recognition task. 
% Finally, we only utilize encoder to fine-tune on the original skeleton sequences for skeleton action recognition.
% Extensive experiments show that our SkeletonMAE achieves remarkable performance for skeleton action recognition in both NTU RGB+D and NTU RGB+D 120 datasets.
Inspired by the MAE \cite{he2022masked}, we propose a spatial-temporal masked autoencoder framework for self-supervised 3D skeleton-based action recognition (SkeletonMAE). Following MAE’s masking and reconstruction pipeline, we utilize a skeleton based encoder-decoder transformer architecture to reconstruct the masked skeleton sequences. A novel masking strategy, named Spatial-Temporal Masking, is 
introduced in terms of both joint-level and frame-level for the skeleton sequence. 
This pre-training strategy makes the encoder output generalizable skeleton features with spatial and temporal dependencies. 
Given the unmasked skeleton sequence, the encoder is fine-tuned for the action recognition task. 
Extensive experiments show that our SkeletonMAE achieves remarkable performance and outperforms %against 
the state-of-the-art methods on both NTU RGB+D and NTU RGB+D 120 datasets.
\end{abstract}

%%%%%%%%% BODY TEXT %%%%%%%%%%

% \begin{figure}[htp]
%   \centering
%   \captionsetup{font=small}
%   \includegraphics[width=0.95\linewidth]{fig_1.pdf}
%   \caption{SkeletonMAE framework. Based on the spatial-temporal masking strategy, a part of the skeleton sequences are masked out and the rest visible sequences are fed to encoder. Then a full set of encoded sequences and masked sequences are processed by decoder for reconstruction. }
%   \label{fig_1}
% \end{figure}

\section{Introduction}
\label{sec:intro}
Human Action Recognition is a fundamental research topic in computer vision, which aims to understand human behaviors and distinguish the actions \cite{sun2022human}. With the booming development of deep learning and human pose estimation methods \cite{cao2017realtime, toshev2014deeppose, sun2019deep}, human skeleton data can be efficiently extracted as a high-level but light-weighted representation, which draws great attention for human behavior and action analysis. Thus, 3D skeleton-based action recognition has become an important research field in human action recognition.

Most recent methods focus on full-supervised learning algorithms to build their frameworks: methods based on Convolutional Neural Network(CNN) \cite{du2015skeleton,zhang2019view}, methods based on Recurrent Neural Networks (RNN) \cite{wang2017modeling,song2017end}, methods based on Graph Convolution Networks (GCN) \cite{yan2018spatial,shi2019two,zhang2020semantics,cheng2020skeleton} and methods based on Transformer \cite{plizzari2021spatial,qiu2022spatio} are widely applied in skeleton action recognition and lead to very good results. 
However, fully supervised action recognition is liable to overfitting. Also, it requires massive labeled training data, which is expensive and time-consuming. To alleviate these issues, \textit{self-supervised learning} methods, which utilize unlabeled data to learn data representations, have been increasingly prevalent in skeleton action recognition.
%There are numerous self-supervised based methods for skeleton action recognition, 
Some self-supervised approaches consider pretext tasks for skeleton representation learning using unlabeled skeleton data, such as motion reconstruction \cite{ cheng2021motion} and jigsaw puzzle \cite{lin2020ms2l}.  However, such pretext-based methods %pay much more attention to 
focus on
local features such as joint correlation and skeleton scale in the same frame, and have not fully explored the  temporal information.
% which aim to build reconstructed skeleton sequences with masked and unmasked skeleton data. Nevertheless, these researches are still relying on the detailed joints information instead of some semantic features (\eg, joint type and frame index \cite{zhang2020semantics}) in time sequence. 
Recently, several works \cite{li20213d, guo2022contrastive} train the contrastive-based model based on contrastive learning framework through constructing the skeleton sequences in different views by data augmentation and positive-negative pairs.
Although these contrastive learning based methods %pay more attention to 
emphasize
high-level context information, they heavily rely on the number of the contrastive pairs in the joints for extracting skeleton features, and ignore the joint correlation information among different frames.

Recently, a new self-supervised learning approach named masked autoencoders (MAE) \cite{he2022masked} demonstrates a strong generalization capability with remarkable performance in computer vision tasks. MAE masks a large proportion of the input image, and then forces the model to learn a generalizable representation by using only the unmasked proportion to reconstruct the original image. However, MAE \cite{he2022masked} can not be directly utilized for self-supervised skeleton action recognition due to the following reasons:
\begin{itemize}[leftmargin=*]
\item The Vision Transformer (ViT) \cite{dosovitskiy2020image} architecture is used in MAE \cite{he2022masked} to process the image input. Different from the image that does not contain temporal information, human skeleton sequences are extracted from videos with high information density, which contains fruitful semantic information: at the spatial level, joint features contain the relationships among different joints in the same frame; in temporal level, frame features represent the movements of the same joint from different frames.
\item The masking strategy in MAE only focuses on the spatial domain. When processing the human skeleton sequences data, a spatial-temporal masking strategy is needed.
\end{itemize}

% Recently, transformer-based network has become one of the most powerful networks in vision tasks via the self-attention mechanism. Due to the strong ability to model global dependencies, transformer architecture is more welcomed than RNNs and GCNs for human pose estimation \cite{zheng2021pose} and skeleton action recognition \cite{plizzari2021spatial} tasks. In the meanwhile, based on the Vision Transformer (ViT), MAE \cite{he2022masked} successfully introduces ViT structures to masked autoencoder, strengthens the capability of learning long-range correspondence in both encoder and decoder. Different from the image-based perspectives above, human skeleton data is extracted from videos with high information density, which contains fruitful semantic information: in spatial level, joint features contains the relationships among different joints in the same frame; in temporal level, frame features represents the movements of the same joint from different frames. Therefore, we are motivated to explore MAE-based self-supervised learning for skeleton action recognition task.  

% As we discussed, the aforementioned self-supervised learning methods pay more attention to the joints features spatially (\eg, the size of hidden pairs in joints \cite{li20213d}), ignoring the effective features in temporal domain and merely utilizing semantic information from skeleton sequences. 
To address these issues, we introduce a novel skeleton-based masked autoencoder named \textbf{SkeletonMAE} for self-supervised skeleton spatial-temporal representation learning: 
1) the masked input sequences are generated from the original skeleton sequences, which contain joints coordinates (spatial) information and frames (temporal) information; 
2) with spatial-temporal masking strategy and encoding-decoding rule, SkeletonMAE gains reconstruction sequences from masked sequences, where the spatial and temporal information is well processed by transformer-based encoder and decoder (transformers have great potential for spatial-temporal representation learning with long-term sequence data).

\begin{figure*}[t]
  \centering
  \captionsetup{font=small}
  \includegraphics[width=0.95\linewidth]{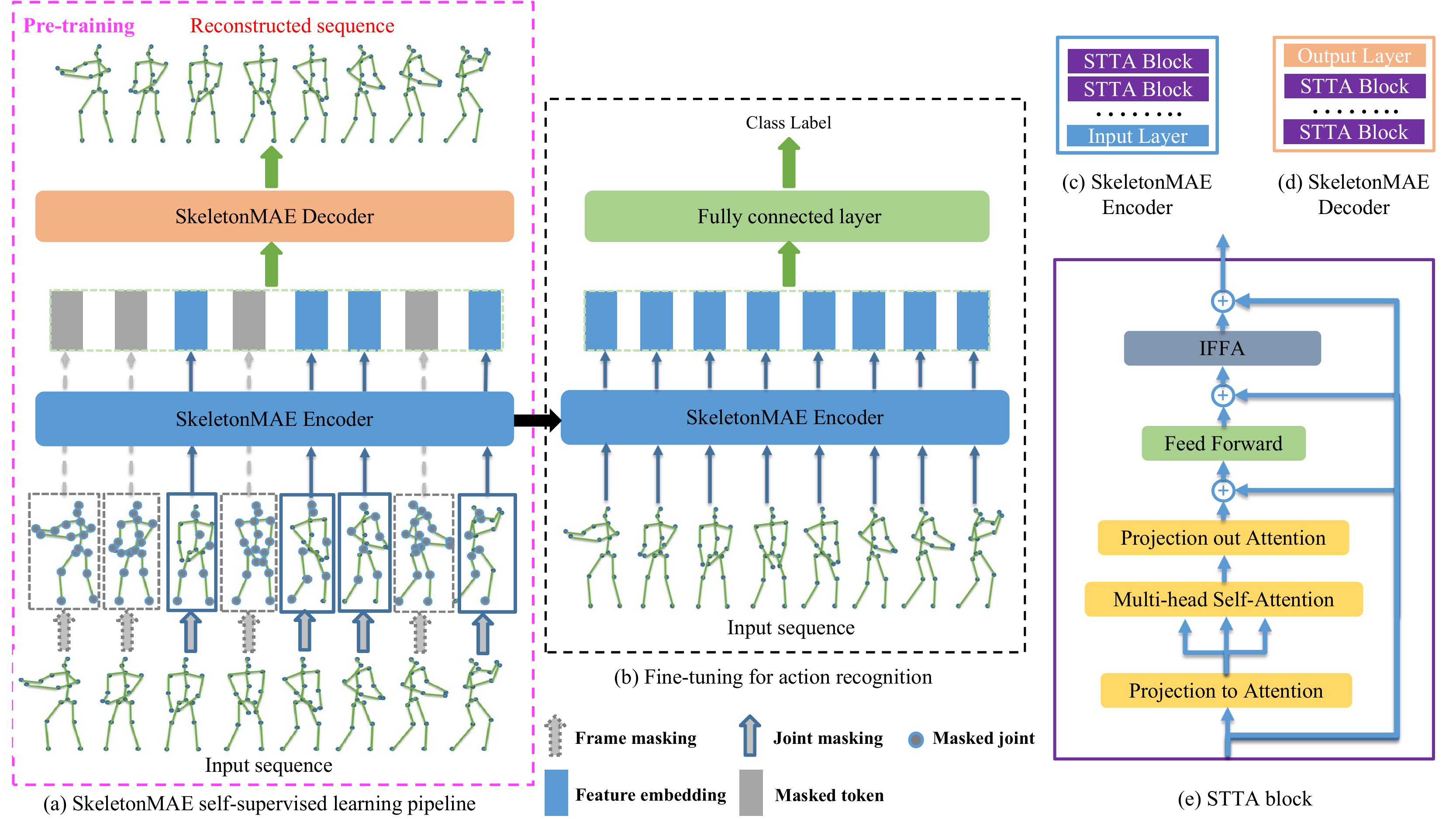}
   \caption{ (a) The overall pipeline of the SkeletonMAE. During the pre-training, we utilize STTFormer to build our encoder and decoder, which consists several STTA blocks respectively. Then we only use SkeletonMAE encoder during the fine-tuning. (b) The end-to-end fine-tuning procedure for skeleton action recognition. (c) The STTFormer-based encoder structure, which is constructed by several Spatio-Temporal Tuples Attention (STTA) blocks and an input layer. (d) The STTFormer-based decoder structure, which is built by a series of STTA blocks with an output layer. (e) The structure of STTA block.}
   \label{fig_2}
\end{figure*}

The framework of SkeletonMAE is presented in Fig. \ref{fig_2}. Specifically, the whole SkeletonMAE pipeline is designed with the following principals. 
During the pre-training stage, 
% The skeleton sequences input contains both frame-level and joint-level information. 
a spatial-temporal masking strategy (with pre-set frame-masking and joint masking ratios) is employed to mask out part of the input skeleton sequence in both frame-level and the joint-level (Sec. \ref{subsec:Spatial-temporal Masking strategies}). In order to find the best trade-off point for spatial-temporal representation learning, we discuss the roles of joint-masking and frame-masking ratios and find the best ratio combination.  
% which produces the masked sequences and keeping the remaining visible sequence. 
The encoder is applied to learn the generalizable feature representation while the decoder is designed to reconstruct the missing skeletons. Since we are dealing with the skeleton sequences, we utilize Spatio-Temporal Tuples Transformer (STTFormer) \cite{qiu2022spatio}, which is developed for processing skeleton sequences, as our network backbone instead of ViT \cite{dosovitskiy2020image}. During the fine-tuning stage, we only use the encoder with a simple output layer to predict the actions. 
The action recognition results show that our approach outperforms the state-of-the-art self-supervised learning methods without extra data. To summarize, we make the following contributions:

\begin{enumerate}[leftmargin=*]
% \item Compared with the traditional self-supervised skeleton action recognition,
\item We propose a simple and efficient skeleton-based masked autoencoder architecture, which aims to learn comprehensive and generalizable skeleton feature representations.
    
\item To have a better understanding of the skeleton masking methods, we explore different masking methods and develop a novel spatial-temporal masking for skeleton data in both joint-level and frame-level. At the same time, we  validate the proper combination of joint-masking ratio and frame-masking ratio.
    
\item We evaluate our model on NTU-RGB+D 60 and NTU-RGB+D 120 datasets, and extensive experimental results show that SkeletonMAE achieves state-of-the-art performance under self-supervised settings.  
\end{enumerate}

\section{Related work}
\label{sec:related_work}
\subsection{\noindent\textbf{Supervised skeleton-based action recognition}} 
In the pre-deep learning period, hand-craft techniques are used for extracting spatial-temporal features in skeleton-based action recognition works \cite{wang2013learning, vemulapalli2014human, vemulapalli2016rolling}. 
In recent years, deep learning has been widely used in skeleton action recognition fields due to its powerful ability of feature extraction and representation learning. And most of them are fully supervised. 
RNN-based methods (\eg, LSTMs) \cite{du2015hierarchical, zhang2017geometric, liu2017skeleton} were widely utilized to process skeleton data. Meanwhile, CNN-based methods \cite{soo2017interpretable, hou2016skeleton, wang2016action} were also introduced to skeleton action recognition. Nevertheless, the data representations extracted by RNNs or CNNs were too simple to present the comprehensive spatial-temporal features of skeleton data. Thus, GCN-based methods \cite{yan2018spatial,shi2019two,zhang2020semantics,cheng2020skeleton} were naturally introduced to model the topological graph features from skeleton date. Recently, with the success of vision transformer (ViT) \cite{dosovitskiy2020image}, transformer-based model becomes powerful architecture for sequential skeleton data analysis \cite{plizzari2021spatial,qiu2022spatio,zheng2021pose,zhang2022mixste,chen2022hierarchically,ce2021gtrs,pang2022igformer} due to the ability of learning global representations. Therefore, we adopt the skeleton-based transformer (STTFormer \cite{qiu2022spatio}) as the backbone network in our research for a better skeleton sequences processing.

\subsection{\noindent\textbf{Self-supervised skeleton-based action recognition}} 
Self-supervised learning aims to extract feature representations without using labeled data, and achieves promising performance in image-based and video-based representation learning \cite{srivastava2015unsupervised, luo2017unsupervised, doersch2015unsupervised, gidaris2018unsupervised}. More self-supervised representation learning approaches adopt the so-called contrastive learning manner \cite{he2020momentum, chen2020simple, li2021contrastive, caron2021emerging} to boost their performance. Inspired by contrastive learning architectures, recent skeleton representation learning works have achieved some inspiring progress in self-supervised skeleton action recognition. MS$^{2}$L \cite{lin2020ms2l} introduced a multi-task self-supervised learning framework for extracting joints representations by using motion prediction and jigsaw puzzle recognition.  CrosSCLR \cite{li20213d} developed a contrastive learning-based framework to learn both single-view and across-view representations from skeleton data. Following CrosSCLR, AimCLR \cite{guo2022contrastive} exploited an extreme data augmentation strategy to add extra hard contrastive pairs, which aims to learn more general representations from skeleton data.

\subsection{\noindent\textbf{Masked 
autoencoding}}
%lu-reword
Masked autoencoding \cite{vincent2008extracting} is a well-structured self-supervised learning model for general representation learning, and successfully applied in BERT \cite{devlin2018bert}, one of the most famous self-supervised frameworks in natural language processing (NLP). The BERT model is simple and straightforward -- remove part of the sequence data with the masked tokens, and predict the removed parts and calculate the loss between prediction and ground-truth data. As a result, the reconstruction sequence works well for training of the generalizable models \cite{liu2019roberta, reimers2019sentence, brown2020language}. Inspired by masked autoencoders and BERT, He et al. \cite{he2022masked}. design a scalable self-supervised masked autoencoder(MAE) for computer vision task. With the same core concept as BERT, MAE masks parts of the image patches and rebuilds them for pre-training. Comparing with the original MAE, there are two main spotlights in our proposed SkeletonMAE: 1) a skeleton-based transformer encoder-decoder framework, the encoder processes the unmasked tokens and decoder reconstructs the original skeleton sequence;
% from a complete patch sequence (assembled by encoded patches and masked patches);  %A high masking ratio (75$\%$ for example)
2) a spatial-temporal masking strategy for both joint and frame level features. Following the main idea of MAE, we propose SkeletonMAE for self-supervised skeleton action recognition.

\begin{figure*}[t]
  \centering
  \captionsetup{font=small}
  \includegraphics[width=0.9\linewidth]{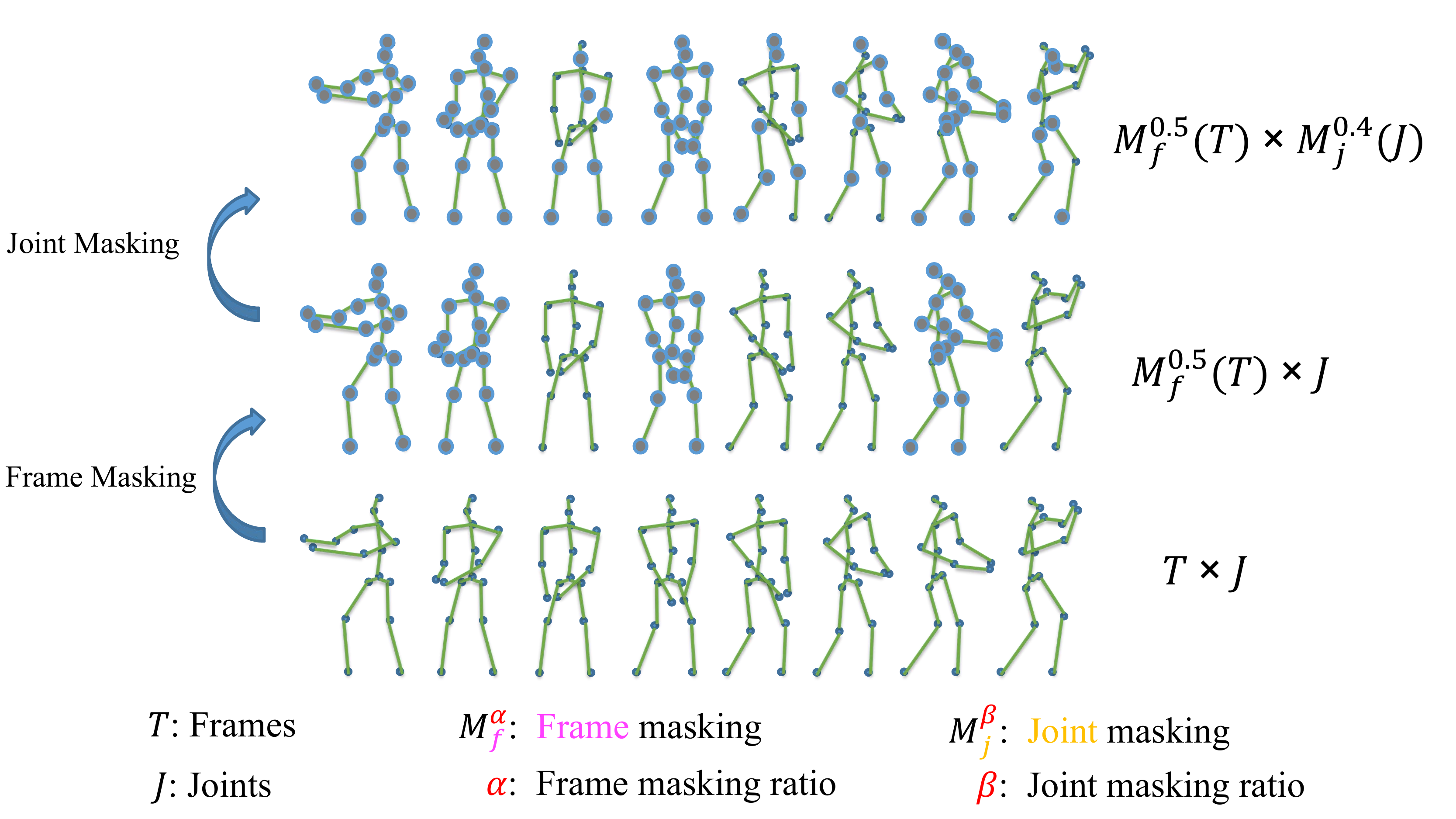}
   \caption{Illustration of the spatial-temporal masking pipeline. Based on pre-set frame-masking ratio ($\alpha$) and joint-masking ratio ($\beta$), we first adopt frame masking (\ie. removing an entire skeleton frame) in skeleton sequence (\eg, $\alpha$ = 0.5), and randomly mask the joints in joint-level (\eg, $\beta$ = 0.4).}
   \label{fig_3}
\end{figure*}

\section{Methodology}
\label{sec:Methodology}
In this section, we first introduce the preliminaries of SkeletonMAE in Sec. \ref{subsec:Preliminaries}. Then, we design a spatial-temporal masking strategy for skeleton data in Sec. \ref{subsec:Spatial-temporal Masking strategies}. Next, we analyze our SkeletonMAE for action recognition in Sec. \ref{subsec:Masked autoencoding architectures}. Finally, we present our fine-tuning procedure in Sec. \ref{subsec:Fine-tuning for skeleton action recognition}.

\subsection{Preliminaries}
\label{subsec:Preliminaries}
\textbf{MAE \cite{he2022masked}.} MAE is formed by an encoder and a decoder in a asymmetric way. It should be noticed that the structure of decoder is different from the encoder, which means that we can adapt some customized decoders to construct the efficient pre-training model. Specifically, the encoder in MAE is based on ViT yet only processing unmasked images: following ViT, the image patches are encoded by linear projection and added with positional embedding to be image tokens, then the tokens are processed by several transformer blocks. Only a small size of the unmasked tokens (75$\%$ patches are masked and the rest are set as the input) are loaded by encoder. As for the MAE decoder, it decodes the masked tokens with the position information based on the original image patches for reconstruction. Then the mean squared error (MSE) is calculated between masked and reconstructed tokens in pixel space. After pre-training, the pre-tained encoder with a simple classification head is applied for image classification task.

\textbf{STTFormer \cite{qiu2022spatio}.} Different from MAE which applies ViT in encoder and transformer blocks in decoder for image reconstruction, we take STTFormer to build encoder and decoder due to its skeleton-based transformer structure. Comparing with ViT which is based on image patches without temporal information, STTFormer is a skeleton data-driven transformer and shows great potential in spatial-temporal data processing. Specifically, STTFormer divides skeleton data into several tuples (non-overlapping parts), and provides the self-attention module named Spatio-Temporal Tuples Attention (STTA) to extract multi-joint representations among adjacent frames. Then a feature aggregation module named Inter-Frame Feature Aggregation (IFFA) is proposed for inter-frame action intergration after STTA block, improving the learning ability for similar action recognition. The structure of STTFormer is shown in Fig. ~\ref{fig_2}.
% Moreover, the encoder in fine-tuning is STTFormer-based as well.

\subsection{Spatial-temporal masking strategy}
\label{subsec:Spatial-temporal Masking strategies}
We propose a spatial-temporal masking method to a portion of the the skeleton sequence input, the pipeline of our masking strategy is illustrated in Fig. \ref{fig_3}.

\textbf{Temporal-masking method.} Fig. \ref{fig_3} shows our masking method at the frame-level. Based on the pre-set frame-masking ratio, a portion of the frames are randomly removed and their indices are stored, the remaining frames are then processed by spatial-masking method at the joint-level.

\begin{figure*}[t]
  \centering
  \captionsetup{font=small}
  \includegraphics[width=0.9\linewidth]{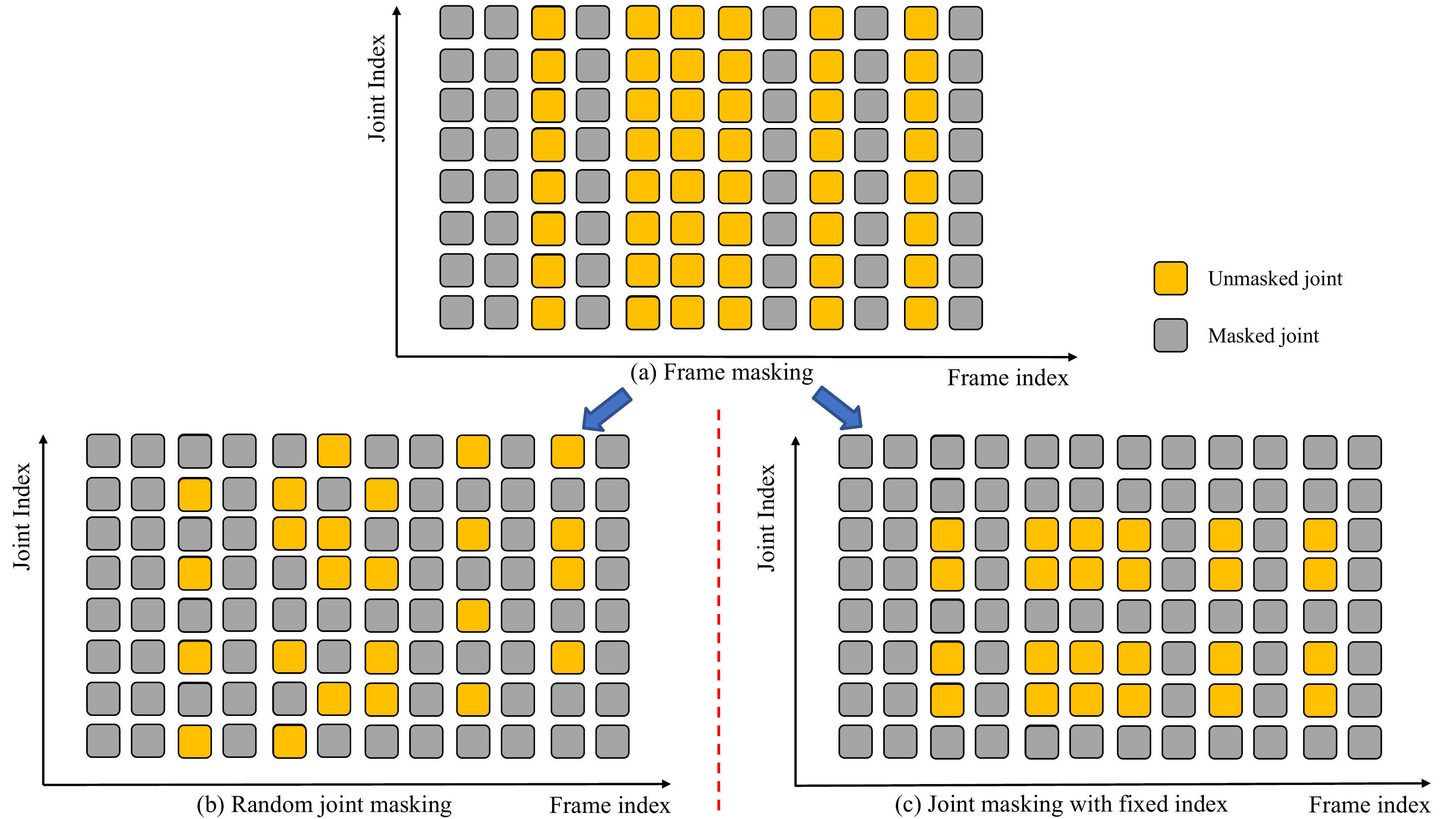}
   \caption{Illustration of two masking strategies. (a) The frame-masking is first implemented and then: (b) randomly mask the joints in spatial level; (b) mask the joint with the fixed index. }
   \label{fig_4}
\end{figure*}

\textbf{Spatial-masking method.} As shown in Fig. \ref{fig_3}, after implementing temporal masking method in all the input frames, the rest frames are then processed via spatial masking strategy. And based on the pre-set joint-masking ratio, we randomly mask part of the joints in every unmasked frame. It is worth noting that the indices of the masked joints are not fixed in this randomly spatial-masking method, which means that the same joints in different frames may be masked or not. This simple approach is illustrated in Fig. \ref{fig_4}(b). Besides this masking method, we also introduce a joint masking strategy with fixed indices, which is shown in Fig. \ref{fig_4}(c). The joints with the same indices in different frames are all masked or not based on the joint-masking ratio. We conduct experiments to compare these two masking strategies in Sec. \ref{subsec:Ablation}.

\subsection{SkeletonMAE architecture}
\label{subsec:Masked autoencoding architectures}
We describe the main components in SkeletonMAE, \eg, encoder, decoder, reconstruction sequence, loss function and fine-tuning pipeline for skeleton action recognition. The pipeline and SkeletonMAE structure are illustrated in Fig. \ref{fig_2}.

\textbf{SkeletonMAE encoder.}
Our encoder is based on STTFormer and only processes the visible skeleton tokens. 
Given a skeleton sequence as input, we apply the frame-masking and joint-masking methods respectively. This spatially and temporally unmasked token is fed to the SkeletonMAE encoder, which maps the input to the spatial-temporal embedding features. 

\textbf{SkeletonMAE decoder.}
Our decoder also adopts STTFormer structure. Same as the decoder in MAE, the spatial-temporal embedding features is processed in SkeletonMAE decoder to reconstruct the original sequence. At the same time, in order to reserve the position information for reconstruction, positional embeddings are also introduced. The output of the decoder is the reconstructed sequence, which should be the same as the original sequence without masking. 

\textbf{Reconstruction.}
We use the mean squared error (MSE) loss to measure the consequence of reconstruction. In this case, we compute the MSE loss between original skeleton sequences and the reconstructed sequences as follows:
\begin{align}
\small
    {MSE} =\frac{1}{N} \sum_{i=1}^N |S_i - S_i^* |^2,
\end{align}
where $i$ is the index of frame,  $N$ is the number of samples, $S$ is the input sequence, and $S^*$ is the reconstructed sequence. 

\subsection{Fine-tuning for skeleton action recognition}
\label{subsec:Fine-tuning for skeleton action recognition}
In order to evaluate SkeletonMAE's ability of learning skeleton representations, we load the learned parameter weights obtained from pre-training to fine-tune the model with all the training data, then the label for each action is predicted with the recognition accuracy. The procedure of fine-tuning is shown in  Fig. \ref{fig_2} (b). Different from the latest contrastive-based self-supervised skeleton action recognition methods\cite{li20213d, guo2022contrastive}, which verify the model via linear evaluation protocol, we only focus on the end-to-end fine-tuning for the skeleton action recognition tasks.

\section{Experiments}
\label{sec:experiments}
% In this section, we first compare our metric with training-free metrics on NAS-Bench-201 using random search and pruning-based search. We then compare our metric with training-free methods and other NAS methods on the DARTS search space. Finally, we conduct ablation study to show the impact of different training hyper-parameters on the result.
% In this section, we first evaluate our method on NAS-Bench-201 and DARTS search spaces. Then we conduct ablation study to show the impact of different training hyper-parameters on the result.

\subsection{Datasets}
\label{sec:datasets}
We evaluate our experiments on the following two most-used datasets: NTU-RGB+D 60 dataset \cite{shahroudy2016ntu} and NTU-RGB+D 120 dataset \cite{liu2019ntu}, and follow the according evaluation protocols for the experimental evaluation.

\textbf{NTU-RGB+D 60 (NTU-60).}
% \label{subsubsec:NTU-60}
NTU-60 is a large scale skeleton dataset for human skeleton-based action recognition, which contains 56,578 videos with 60 action categories and each human body contains 25 joints. There are two evaluation protocols for NTU-60: Cross-Subject (X-Sub) and Cross-View (X-View) protocols. X-Sub protocol means training data and validation data are split by different subjects, and half of the subjects are set as training sets and the rest are the test sets. X-View protocol means training data and validation data are collected from different camera views (camera 1,2 and 3). In X-View, the samples captured by camera 2 and 3 are set for training and the samples of camera 1 are set as testing set.

\textbf{NTU-RGB+D 120 (NTU-120).}
% \label{subsubsec:NTU-120}
NTU-120 is an expansion dataset of NTU-60 with 113,945 sequences with 120 action labels. There are also two evaluation protocols: Cross-Subject (X-Sub) and Corss-Set (X-Set). In X-Sub, there are 53 subjects for training and 53 subjects for testing. In X-Set, half of the setups are split for training (even setup IDs) and the rest (odd setup IDs) are used for testing.

\subsection{ Experimental settings}
\label{subsec:Experimental settings}
Our experiments are performed on 8$\times$ A6000 GPUs with Pytorch \cite{paszke2019pytorch} framework implementation. Both our pre-training and fine-tuning models are trained by Adam optimizer \cite{kingma2014adam} with base learning rate 0.005 and weight decay 0.0001. The batch size is 64. The pre-training and fine-tuning epoch number are all set to 200. We also use a multi-step learning rate schedule for learning rate adjustment with gamma 0.1 and milestones are 60 epoch, 90 epoch and 110 epoch.
For fair comparisons among different methods, we limit the length of the skeleton sequence to 20 frames for all experiments.

\textbf{STTFormer.}
% \label{subsubsec:STTFormer settings}
As mentioned in Sec \ref{subsec:Preliminaries}, in order to learn better spatial-temporal representations, we utilize the standard STTFormer as our backbone in pre-training, which consists of a stack of STTA Blocks. Same as MAE, our method also adds sine-cosine positional embedding to both encoder and decoder inputs. For fine-tuning, we use the STTFormer-based encoder as our feature extractor.

\textbf{Sequence division and patch embedding.}
In our research, we follow the patch embedding method in STTFormer. We first divide the original skeleton sequence into tuples. Then, since the skeleton data does not contain a large number of pixels and various noises like image data, we directly use a 1x1 Conv for patch embedding processing.

\textbf{Masking settings.}
% \label{subsubsec:Masking strategies}
We implement our masking strategies before sequence division. As we discussed in Sec. \ref{subsec:Spatial-temporal Masking strategies},  we first mask out a random subset of frames by the pre-set frame masking ratio and then mask out a random index of joints by the pre-set joint masking ratio. During experiments, we test several trials of frame-masking ratio and joint-masking ratio, finding the best trade-off combination.

\textbf{Pre-training.}
We choose MSE loss as pre-training loss and save the best model by the minimized validation loss.

\textbf{Fine-tuning.}
As we discussed in Sec. \ref{subsec:Fine-tuning for skeleton action recognition}, we use end-to-end fine-tuning for the end task. Moreover, we choose cross entropy loss with label smoothing \cite{he2019bag} as fine-tuning loss with smoothing rate 0.1 and save the best model by the maximized validation accuracy.

\subsection{Ablation study}
\label{subsec:Ablation}
\textbf{Different masking strategies.}
After performing the same degree of random frame masking, we compare the masking strategies of masking the joints randomly with the method keeping the same masked joint index over the entire sequence. The experimental results show that the pure random joint masking for visible frames is more helpful for the final fine-tuning result (in Table \ref{tab: table1}, we get our best fine-tuned recognition accuracy of $86.6\%$ on X-sub using random masking strategy, which is $1.2\%$ better than the best result of the masking method by fixing the joints indices). The overall results indicate that the random masking method outperforms the masking method with fixed joints indices, which means the model learns better features with a randomly generated input than the pre-defied input. Notably, MAE experiment also shows that using a more random masking strategy is more beneficial to the final fine-tuning result.

% Therefore, designing an optimal masking strategy depends on the different characteristics from different data.
% However, VideoMAE states that tube masking considering the temporal dimension of video data is better than random sampling.

\textbf{Frame-masking ratio and joint-masking ratio.}
In spatial and temporal domains, we test several combinations of different frame-masking ratio and joint-masking ratio on SkeletonMAE. Following both joint index fixed and random masking strategies, we set the frame masking ratio 0.4, 0.5 and 0.6 respectively, for every decided frame masking ratio, we test different joint masking ratio (0.4, 0.5 and 0.6 respectively). As shown in Table \ref{tab: table1}, the final results on NTU-60 with X-Sub show that a frame-masking ratio of 0.4 and a joint-masking ratio of 0.5 work best in the masking method with fixed joints indices ($85.4\%$ accuracy). 
Using the random masking method, we achieve the best result ($86.6\%$ accuracy) in two combinations (0.5 joint-masking ratio with 0.5 or 0.4 frame-masking ratio). 

%%%%%%%%%%%%%%%%%%%%%%%%%%%%%%%%%%%%%%%%%%%%%%%%%%%

\begin{table}[htp]
% \tiny
% \scriptsize
% \renewcommand\arraystretch{1.2}
\vspace{-5pt}
\centering
  \resizebox{1\linewidth}{!}
  { 
\begin{tabular}{c|cc|c}
\hline
method                   & frame-masking ratio & joint-masking ratio  &  NTU-60 X-Sub \\ \hline
                         & 0.6           & 0.4                               & 85.2                                 \\
                         & 0.6           & 0.5                                & 84.9                                 \\
                         & 0.6           & 0.6                               & 85.3                                 \\
                         & 0.5           & 0.4                              & 85.3                                 \\
                         & 0.5           & 0.5                              & 85.0                                   \\
                         & 0.5           & 0.6                               & 84.8                                 \\
                         & 0.4           & 0.4                               & 84.8                                  \\
                          & 0.4           & 0.5                               & \textbf{85.4}                        \\
\multirow{-9}{*}{fixed index} & 0.4           & 0.6                              & 85.2                                 \\ \hline
                         & 0.6           & 0.4                               &  86.5                                \\
                         & 0.6           & 0.5                                & 86.0                                 \\
                         & 0.6           & 0.6                               & 86.3                                 \\
                         & 0.5           & 0.4                              & 86.3                                 \\
                         & 0.5           & 0.5                              & \textbf{86.6}                                  \\
                         & 0.5           & 0.6                               & 85.7                                 \\
                         & 0.4           & 0.4                               & 85.6                                 \\
                         & 0.4           & 0.5                               & \textbf{86.6} \\
\multirow{-9}{*}{random} & 0.4           & 0.6                              & 85.4                                 \\ \hline
\end{tabular}
}
\caption{Masking strategies with joint-masking ratio and frame-masking ratio. Specifically, there are two joint masking methods tested: fixed indices masking and randomly masking. }
\label{tab: table1}
\vspace{-5pt}
\end{table}

%%%%%%%%%%%%%%%%%%%%%%%%%%%%%%%%%%%%%%%%%%%%%%%%%%

\textbf{Embedding dimension.}
Table \ref{tab: table2} shows the ablation study on the embedding dimension of the decoder. We change the different embedding dimensions in SkeletonMAE decoder and find that the default setting with 256 dimension works better (86.6$\%$ accuracy) than the larger size (86.0$\%$ accuracy) and the small size (85.2$\%$ accuracy). We also observe that with the increasing size of embedding dimension, the number of model parameters increase as well, when we set the dimension as 512, the parameters are 11 times larger than the parameters with dimension 128, which costs more time for training. So we choose 256 as the default embedding dimension for the following ablation studies.

%%%%%%%%%%%%%%%%%%%%%%%%%%%%%%%%%%%

\begin{table}[htp]
% \tiny
% \scriptsize
% \renewcommand\arraystretch{1.2}
\vspace{-5pt}
\centering
  \resizebox{0.8\linewidth}{!}
  {
\begin{tabular}{c|c|c}
\hline
embedding dimension & NTU-60 X-Sub                          & parameters(M) \\ \hline
128                 &85.2                                     &3           \\ 
256                 &\textbf{86.6}    &11            \\ 
512                &86.0                                       &33            \\ \hline
\end{tabular}
}
 \caption{Ablation study on embedding dimension.}
\label{tab: table2}
\vspace{-4pt}
\end{table}
%%%%%%%%%%%%%%%%%%%%%%%%%%%%%%%%%%%%%%%%%%%%%%%%%%%%%%

\textbf{Decoder depth.}
Decoder depth represents the number of the STTFormer blocks. According to the last ablation experiment, we set the embedding dimension (the width of the decoder) as the default size 256, and vary the decoder depth (11, 9, 7 and 5 blocks). As the results shown in Table \ref{tab: table3}, SkeletonMAE achieves the best result (86.6$\%$ accuracy) when the decoder depth is 9. The deep depth (11 blocks with 86.5$\%$ accuracy) and shallow depth (7 blocks with 86.2$\%$ accuracy and 5 blocks with 85.7$\%$ accuracy) perform worse. According to the results from embedding dimension and decoder depth experiments, we finalize our default decoder configurations for the following experiments (256 embedding dimension and 9 blocks).

%%%%%%%%%%%%%%%%%%%%%%%%%%%%%%%%%%%%%%%%%%
\begin{table}[htp]
% \tiny
% \scriptsize
% \renewcommand\arraystretch{1.2}
\vspace{-4pt}
\centering
  \resizebox{0.6\linewidth}{!}
  {
\begin{tabular}{c|c}
\hline
decoder depth & NTU 60 X-Sub                             \\ \hline
11            & 86.5                                     \\ 
9             & \textbf{86.6}                                    \\ 
7             & 86.2                           \\ 
5             & 85.7                                     \\ \hline
\end{tabular}
}
\caption{Ablation study on decoder depth.}
\label{tab: table3}
\vspace{-4pt}
\end{table}
%%%%%%%%%%%%%%%%%%%%%%%%%%%%%%%%%%%%%%%%%%%%%%%

\begin{figure}[thp]
  \centering
  \captionsetup{font=small}
  \includegraphics[width=1\linewidth]{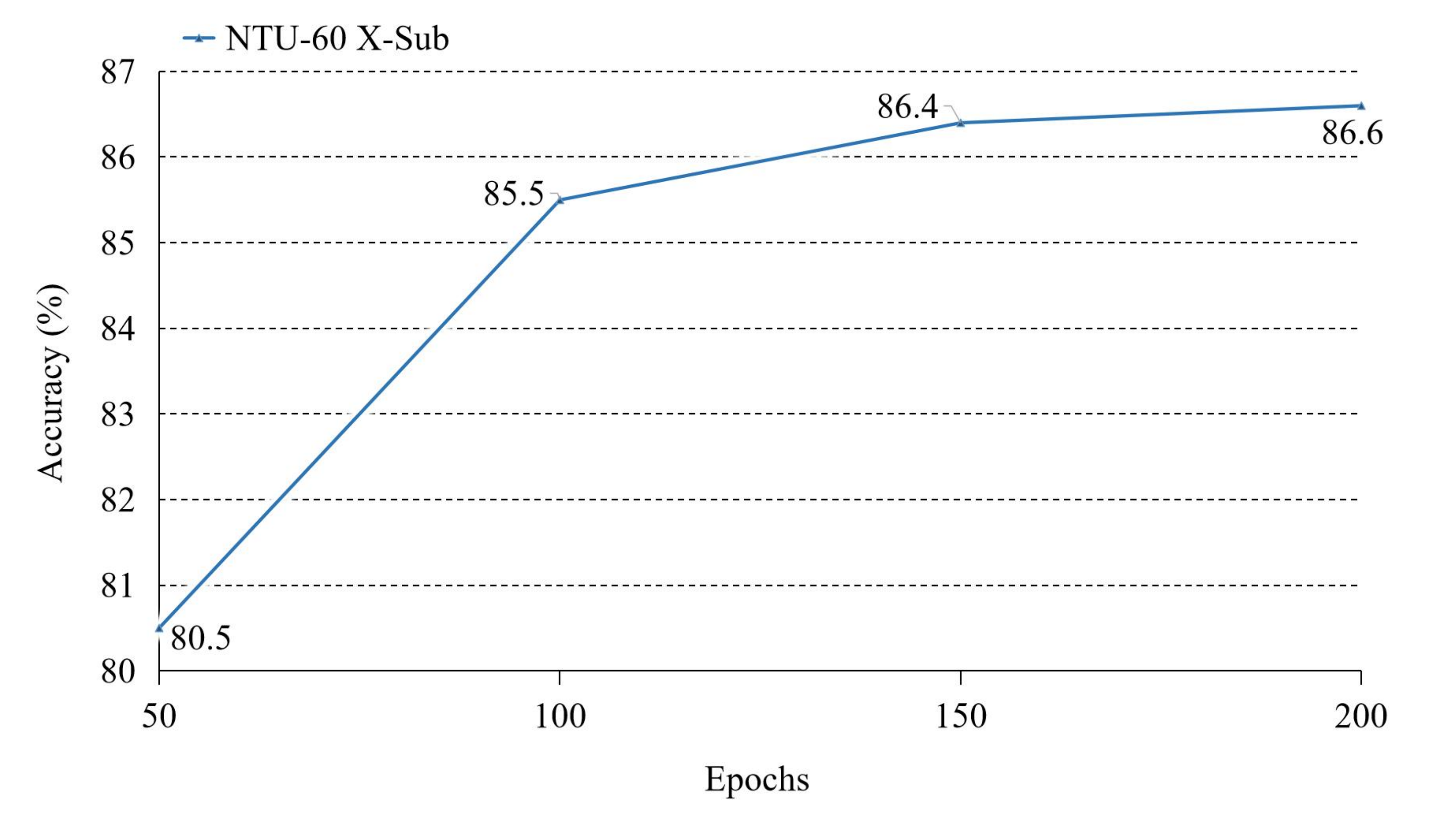}
   \caption{Ablation study on pre-training schedule. }
   \label{fig_5}
\end{figure}

\textbf{Pre-training schedule.}
Normally,  a longer pre-training schedule will give an improvement, thus in this ablation study, we increase the pre-training epoch from 50 epoch to 200 epoch, and test the best fine-tuned results at every 50 epoch. As it shown in Fig. \ref{fig_5} , the best accuracy is 86.6$\%$, so we select 200 epoch as the default pre-training epoch for the following experiments. It is worth noting that there is an impressive improvement (5.0$\%$) between 50 epoch to 100 epoch, but a slight improvement (0.2$\%$) between 150 epoch to 200 epoch, which means it is not cost-effective to keep increasing the pre-training epoch.

%%%%%%%%%%%%%%%%%%%%%%%%%%%%%%%%%%%%%%%%%%%%%%%%%%%%
% \begin{table}[htp]
% % \tiny
% % \scriptsize
% % \renewcommand\arraystretch{1.2}
% \vspace{-4pt}
% \centering
%   \resizebox{0.6\linewidth}{!}
%   {
% \begin{tabular}{c|c}
% \hline
% number of epoch & NTU-60 X-Sub                       \\ \hline
% 50              & 80.1                                 \\ 
% 100             & 85.5                                 \\ 
% 150             & 86.4                                 \\ 
% 200             & 86.6                         \\ \hline
% \end{tabular}
% }
%   \caption{Ablation study on pre-training schedule.}
% \label{tab: table4}
% \vspace{-5pt}
% \end{table}
%%%%%%%%%%%%%%%%%%%%%%%%%%%%%%%%%%%
%%%%%%%%%%%%%%%%%%%%%%%%%%%%%%%%%%%%%%%%%%%%%%%%%%%%

\subsection{Comparison with state-of-the-art}
\textbf{Self-supervised training.}
Notably, as we can see from Table \ref{tab: table5}, our SkeletonMAE outperforms two latest self-supervised skeleton action recognition methods: CrosSCLR \cite{li20213d} and AimClR \cite{guo2022contrastive}. For a fair comparison, we replace their backbone networks (both of them use ST-GCN as the backbone) with STTFormer under the same settings. The results show  that on NTU-60 dataset, our SkeletonMAE leads CrosSCLR $2.0\%$ and AimCLR $2.7\%$ on X-Sub, and also leads CrosSCLR $2.4\%$ and AimCLR $2.5\%$ under X-View protocol. As for the results on NTU-120 dataset, SkeletonMAE outperforms CrosSCLR by $1.8\%$ and $1.2\%$ on X-Sub and X-Set, and also outperforms AimCLR by $2.2\%$ and $1.9\%$ on X-Sub and X-Set respectively. The results indicate that our SkeletonMAE not only achieves outperforming results on the small-size dataset but also the large-size dataset.

%%%%%%%%%%%%%%%%%%%%%%%%
\begin{table}[htp]
% \tiny
% \scriptsize
% \renewcommand\arraystretch{1.2}
\vspace{-5pt}
\centering
  \resizebox{1\linewidth}{!}
  {
\begin{tabular}{c|c|cc|cc}
\hline
                                      &              & \multicolumn{2}{c|}{NTU-60}                                                 & \multicolumn{2}{c}{NTU-120}                       \\ \hline
method                                & backbone & X-Sub                                & X-View                               & X-Sub       & X-Set                               \\ 
\hline
CrosSCLR\cite{li20213d}                              & ST-GCN       & 82.2                                 & 88.9                                 & 73.6        & 75.3                                \\ 
AimCLR\cite{guo2022contrastive}                                  & ST-GCN       & 83.0                                   & 89.2                                 & 76.4        & 76.7                                \\ \hline 
CrosSCLR\cite{li20213d}                              & STTFormer          & 84.6                                 & 90.5                                 & 75.0 & 77.9                         \\ 
AimCLR\cite{guo2022contrastive}                                  & STTFormer          & 83.9                                 & 90.4                                 & 74.6 & 77.2 \\ 
SkeletonMAE & STTFormer          & {\textbf{86.6}} & { \textbf{92.9}} &{\textbf{76.8}}  & {\textbf{79.1}}                           \\ \hline
\end{tabular}
}
  \caption{Fine-tuned results on NTU-60 and NTU-120 datasets.}
\label{tab: table5}
\vspace{-5pt}
\end{table}
%%%%%%%%%%%%%%%%%%%%%%%%%%%%%%%%%%%%%%%%%%%%%%%%
\begin{figure*}[!ht]
\vspace{-0.5cm}
  \centering
  \captionsetup{font=small}
  \includegraphics[width=\linewidth, height=0.55\linewidth]{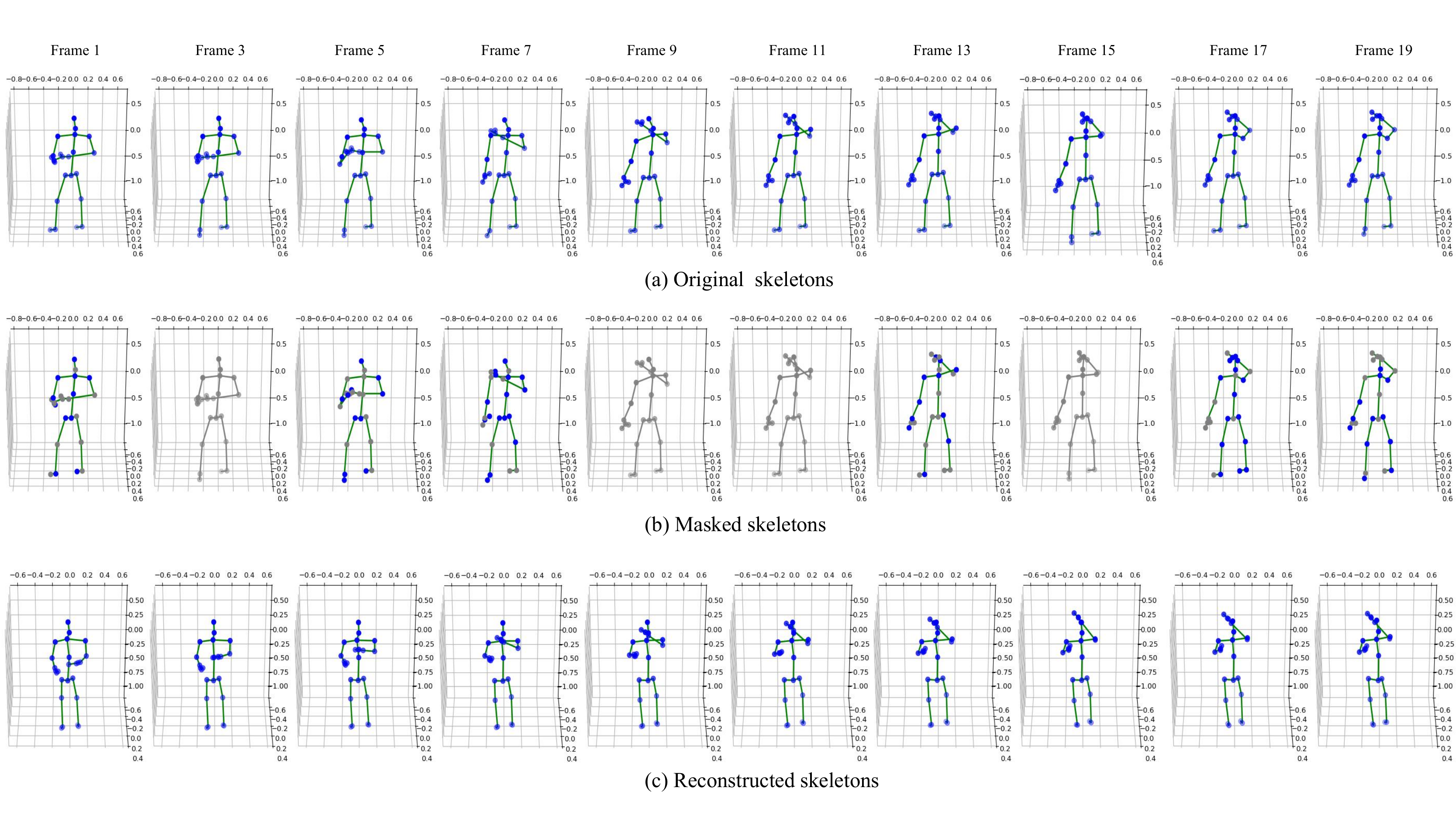}
  \vspace{-0.75cm}
   \caption{Visualization results on NTU-60 dataset, \textit{drink water} action. We select the odd frames from the first 20 frames, for each frame, we visualize: (a) the input skeleton data; (b) the masked skeleton data (0.5 joint-masking ratio); (c) the reconstructed skeleton frames .}
   \label{ske_1}
\end{figure*}

%%%%%%%%%%%%%%%%%%%%%%%%%%%%%%%%%%%%%%%%%%%%%%%%

\textbf{Fewer labeled data training.}
In order to figure out the ability of spatial-temporal feature learning in fewer-data situation, we fine-tune our pre-trained SkeletonMAE model with only $5\%$ and $10\% $ labeled data on both NTU-60 and NTU-120 datasets. According to Table \ref{tab: table6}, our SkeletonMAE achieves $64.4\%$ and $68.8\%$ on NTU-60 X-Sub and X-View with only $5\%$ fine-tuning data, and surpasses CrossSCLR and AimCLR. Moreover, our SkeletonMAE also performs better than CrossSCLR and AimCLR with $10\% $ labeled data ($73.0\%$ and $76.9\%$ 
on NTU-60 X-Sub and X-View respectively). Meanwhile, our SkeletonMAE achieves outperformed results on NTU-120 data with $5\%$ ($50.4\%$ on X-Sub and $52.0\%$ on X-Set) and $10\%$ ($61.8\%$ on X-Sub and $62.5\%$ on X-Set) labeled data, 
which demonstrates a better capability of generalizability learning of our approach under the extreme fine-tuning situation.

%%%%%%%%%%%%%%%%%%%%%%%%%%%%%%%%%%%%
\begin{table}[htp]
% \tiny
% \scriptsize
% \renewcommand\arraystretch{1.2}
\vspace{-5pt}
\centering
  \resizebox{1\linewidth}{!}
  {
\begin{tabular}{c|c|c|cc|cc}
\hline
\multirow{2}{*}{method} & \multirow{2}{*}{backbone} & \multirow{2}{*}{label fraction} & \multicolumn{2}{c}{NTU-60} & \multicolumn{2}{c}{NTU-120}    \\ \cline{4-7} 
                        &                           &                                 & X-Sub         & X-View        & X-Sub         & X-Set\\ \hline
CrosSCLR\cite{li20213d}             & STTFormer                       & 5\%                             & 63.5          & 66.9         & 50.2         & 50.4 \\
AimCLR\cite{guo2022contrastive}                    & STTFormer                       & 5\%                             & 63.9          & 67.5         & 49.0         & 51.8 \\
SkeletonMAE             & STTFormer                       & 5\%                             & \textbf{64.4} & \textbf{68.8}  &   \textbf{50.4}          & \textbf{52.0} \\ \hline
CrosSCLR\cite{li20213d}             & STTFormer                       & 10\%                            & 71.0            & 75.1          & 58.5         & 60.6\\
AimCLR\cite{guo2022contrastive}                  & STTFormer                       & 10\%                            & 70.2          & 76.2          & 58.6         & 60.5\\
SkeletonMAE             & STTFormer                       & 10\%                            & \textbf{73.0}   & \textbf{76.9}  & \textbf{61.8}         & \textbf{62.5} \\ \hline
\end{tabular}
}
  \caption{Fine-tuned results with fewer labeled data on NTU-60 and NTU-120 datasets.}
\label{tab: table6}
\vspace{-5pt}
\end{table}

%%%%%%%%%%%%%%%%%%%%%%%%%%%

%%%%%%%%%%%%%%%%%%%%%%%%%%%%%%%%%%%%%%%%%%%%%%%%%%%

\subsection{Visualization}
\label{subsec:visualization}

In Fig. \ref{ske_1}, we show the visualization results of SkeletonMAE pre-training on NTU-60 dataset using randomly joint masking strategy with 0.5 joint masking ratio (a frame-masking is applied first). We select the odd frames from the first 20 frames from the \textit{drink water} action. As it shown: Fig. \ref{ske_1} (a) visualizes the input skeleton, Fig. \ref{ske_1} (b) shows the corresponding masking results in both frame-level (frame 3, 9, 11, 15) and joint-level (frame 1, 5, 7, 13, 17, 19), Fig. \ref{ske_1} (c) 
shows the reconstructed skeletons of the pre-training.
Spatially (the visualization results in the same frame), we observe that there exist a few detailed differences between the original skeleton sequence and the reconstructed skeleton sequence, but the frameworks of the human body (\eg, the positions of arms and legs) are kept without distortion. The detail difference visualization shows the good ability of SkeletonMAE for spatial-feature learning. 
Temporally (the consecutive skeleton sequences), although we also observe a few variations between the original sequence and the reconstructed ones, there is no pronounced deformation in the time space (the joint motion in different frames is reserved, \eg, rising hands), which indicates that our SkeletonMAE learns temporal representations well. The overall results demonstrate that SkeletonMAE learns generalized skeleton sequences containing semantic action information, resulting a good performance in action recognition task.

\section{Conclusion}
\label{sec:conclusion}
We conduct a novel skeleton-based masked autoencoder named SkeletonMAE for self-supervised skeleton action recognition. In order to get a better skeleton representation learning, we apply a novel spatial-temporal masking strategy in pre-training for skeleton reconstruction. The roles of different frame-ratio and joint-ratio are also discussed and implemented. With comprehensive experiments on NTU-60 and NTU-120 datasets, we show outperformed results of SkeletonMAE for skeleton action recognition.

{\small
\bibliographystyle{ieee_fullname}
\bibliography{egbib}
}

%%%%%%%%%%%%%%%%%%%%%%%%%%%%%%%%%%%%%%%%%%%%%%%%%%%%%%%%%
\clearpage
\appendix
\section*{Supplementary material}
In this supplementary material, we provide the following items for better understanding the paper:
\begin{itemize}[leftmargin=*]
\item Detailed architectures of SkeletonMAE encoder and decoder.
\item More visualization results.
\item Qualitative analysis on masking strategy.
\end{itemize}

% \title{SkeletonMAE: Spatial-Temporal Masked Autoencoders for Self-supervised
% Skeleton Action Recognition - Supplementary Material }

% \and

\maketitle
\thispagestyle{empty}

\subsection*{A. Detailed architecture}
\label{sec:detailed architecture}
As it shown in Table \ref{tab: table1}, we give the detailed architecture of SkeletonMAE, including the dimensions of input and output layers, the size of each STTA block (including the input dimension $D_{in}$ and output dimension $D_{out}$) in the encoder and decoder. 

Given the original 3D skeleton data: \begin{equation} X_{ori}\in R^{T\times J\times D}, \end{equation} where $T$ is the number of frames of the input sequence, $J$ is the number of  joints in each frame, $D$ is the input dimension ($J$ = 25 and $D$ = 3 in NTU-60 and NTU-120 datasets). Then based on the spatial-temporal masking 
strategy we proposed, after applying a masking method $M$ to $X_{ori}$, the input skeleton data is expressed as: \begin{equation} X_{in} = M(X_{ori}), \end{equation}  \begin{equation} X_{in}\in R^{T'\times J'\times D_{in}}, \end{equation} where $T'$ and $J'$ represent the masked frames and masked joints following the pre-set masking approach. In SkeletonMAE encoder, according to the structure of STTA block and the data processing from STTFormer, $X_{in}$ is processed by a series of STTA blocks for data embedding:
 \begin{equation} X_{out} = STTA_i(x_{in}), i \in [1,...,N], \end{equation} 
 \begin{equation} X_{out}\in R^{T'\times J'\times D_{out}}\end{equation} 
 where $N$ is the number of STTA blocks and $D_{out}$ is the output dimension ($D_{out}$ = 3). As for the decoder, it has an inverted structure of the decoder (the output layer is at the end of the decoder).  

Moreover, we provide the sizes of query \textbf{Q}, key \textbf{K} and value \textbf{V} from the STTA blocks. Specifically, the embedding dimension for Table \ref{tab: table1} is 256 and the number of blocks in both encoder and decoder is 9 (the default setting for ablation studies). Finally, we also give more settings of the ablation studies on embedding dimension and decoder depth (Table \ref{tab: table2}, \ref{tab: table3}, \ref{tab: table4}, \ref{tab: table5}, \ref{tab: table6}).

%%%%%%%%%%%%%%%%%%%%%%%%%%%%%%%%%%%%%%%%%%%%%%%%
\subsection*{B. More visualization}
\label{sec:more visualization}
\begin{figure*}[htp]
\vspace{-0.5cm}
  \centering
  \captionsetup{font=small}
  \includegraphics[width=\linewidth, height=0.55\linewidth]{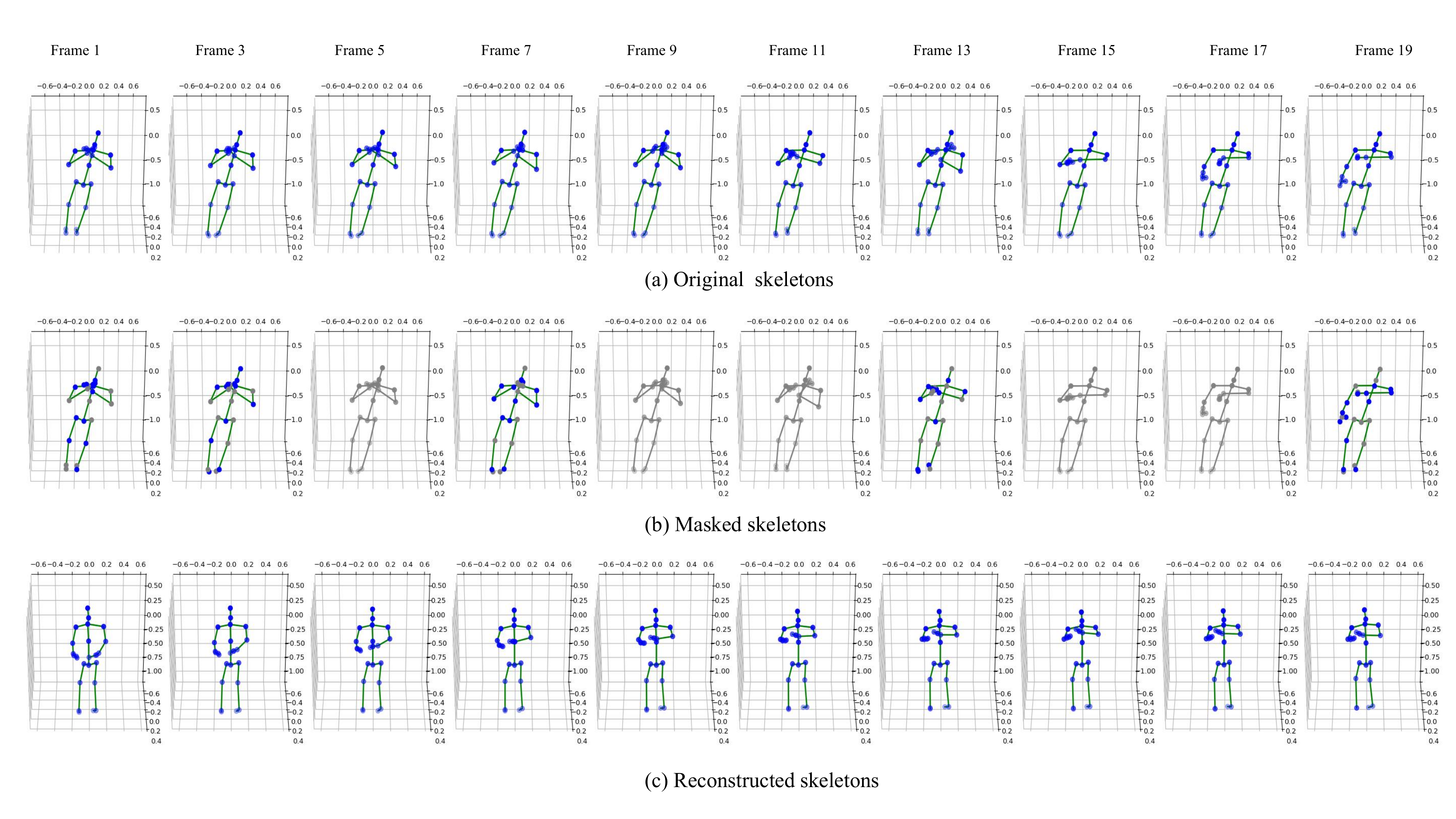}
  \vspace{-0.75cm}
   \caption{Visualization results on NTU-60 dataset, \textit{type on a keyboard} action.}
   \label{ske_2}
\end{figure*}

\begin{figure*}[htp]
\vspace{-0.5cm}
  \centering
  \captionsetup{font=small}
  \includegraphics[width=\linewidth, height=0.55\linewidth]{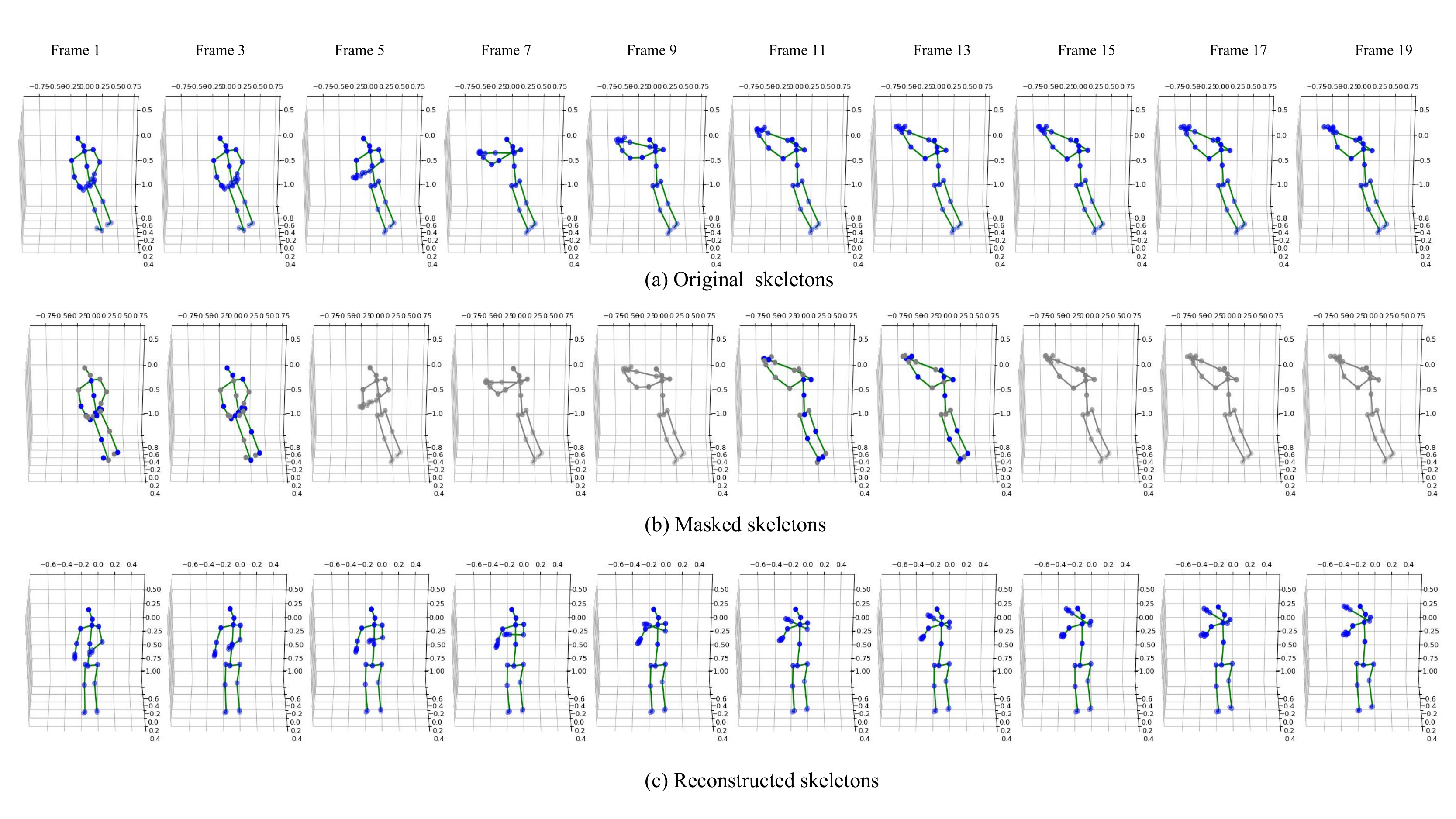}
  \vspace{-0.75cm}
   \caption{Visualization results on NTU-60 dataset, \textit{taking a selfie} action.}
   \label{ske_3}
\end{figure*}

\begin{figure*}[htp]
\vspace{-0.5cm}
  \centering
  \captionsetup{font=small}
  \includegraphics[width=\linewidth, height=0.55\linewidth]{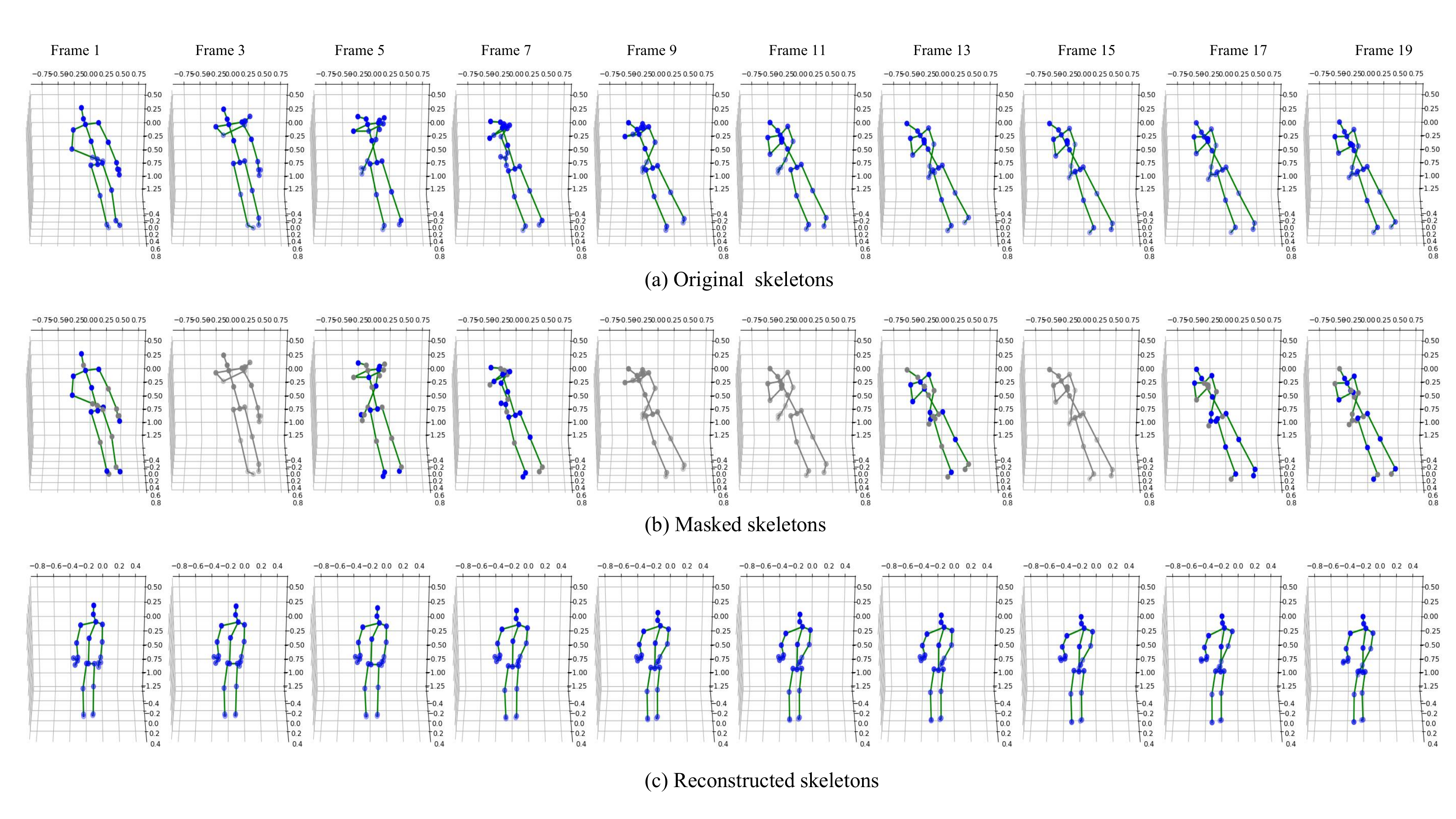}
  \vspace{-0.75cm}
   \caption{Visualization results on NTU-60 dataset, \textit{back pain} action.}
   \label{ske_4}
\end{figure*}

\begin{figure*}[htp]
\vspace{-0.5cm}
  \centering
  \captionsetup{font=small}
  \includegraphics[width=\linewidth, height=0.55\linewidth]{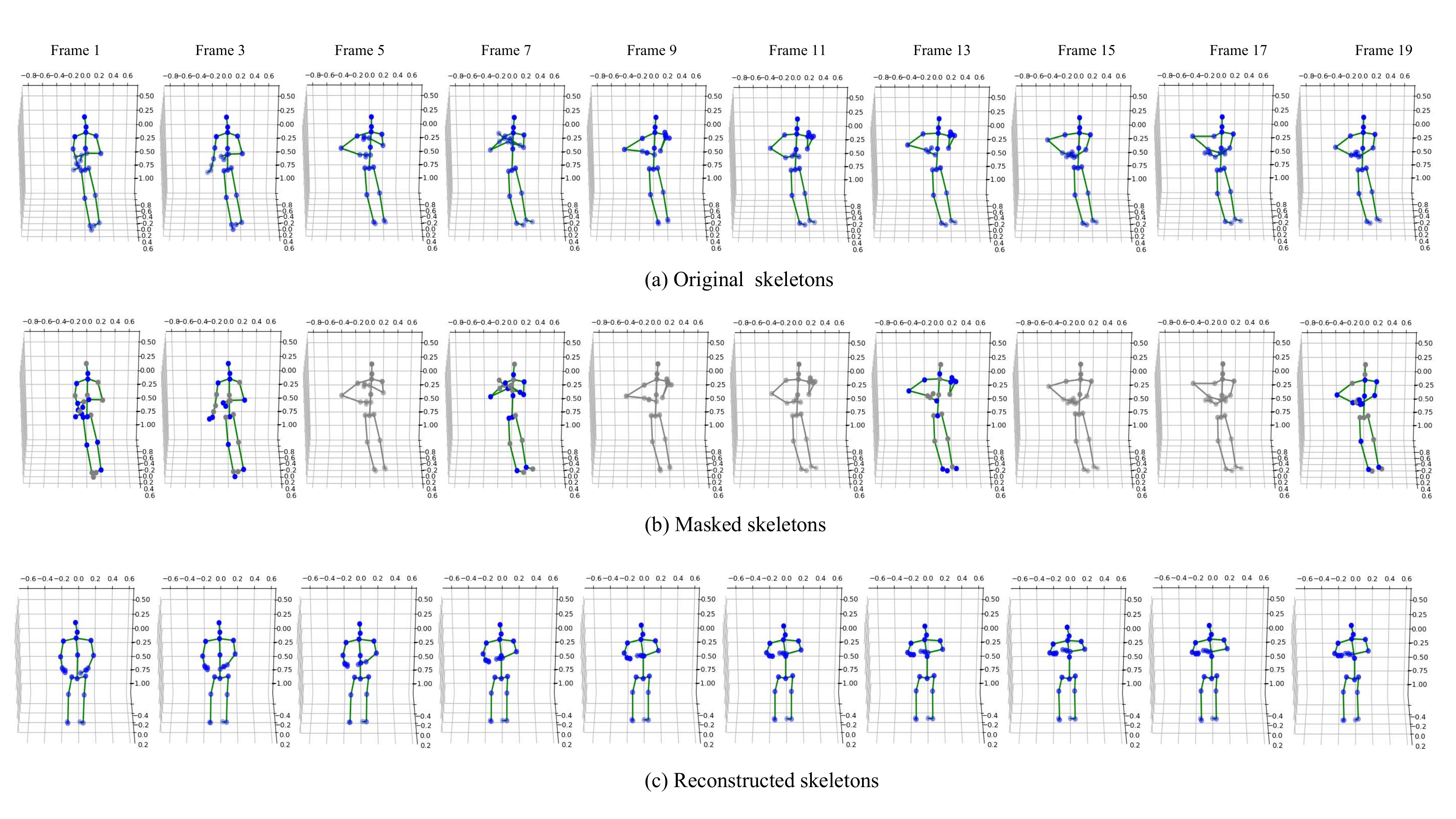}
  \vspace{-0.75cm}
   \caption{Visualization results on NTU-60 dataset, \textit{fan self} action.}
   \label{ske_5}
\end{figure*}

We provide more visualization results of SkeletonMAE pre-training on NTU-60 dataset with random joint-masking method (joint-masking ratio is 0.5) in Fig. \ref{ske_2}: \textit{type on a keyboard}; Fig. \ref{ske_3}: \textit{taking a selfie}; Fig. \ref{ske_4}: \textit{back pain};  Fig. \ref{ske_5}: \textit{fan self}. As we discussed before, the SkeletonMAE learns generalizable features from skeleton data without pronounced deformation. However, the detailed differences between the original skeleton and the reconstructed skeleton in the same frame are still observed (\eg, the coordinates of the forearms from the reconstructed skeletons are different from the original skeletons).
%%%%%%%%%%%%%%%%%%%%%%%%%%%%%%%%%%%%%%%%%%%%%%

\subsection*{C. Qualitative analysis on masking strategy}
\label{sec: Qualitative}
\begin{figure}[h]
\vspace{-11pt}
  \centering
  \captionsetup{font=small}
  \includegraphics[width=1\linewidth]{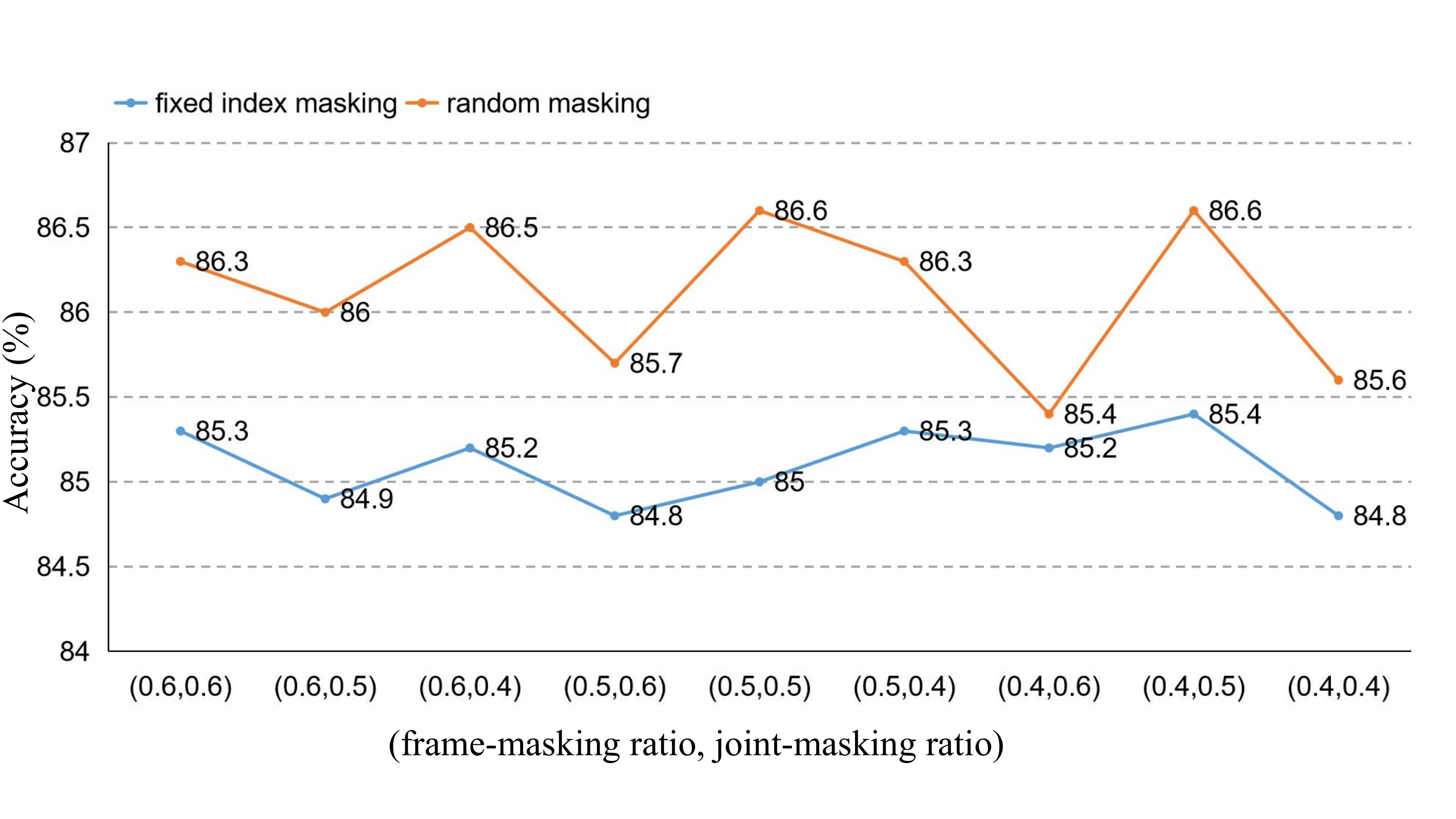}
   \caption{Qualitative analysis on different masking strategies (NTU-60 dataset X-Sub). }
   \label{fig_7}
\vspace{-15pt}
\end{figure}

In Fig. \ref{fig_7}, we show the qualitative analysis results of two masking strategies (index fixed and random masking methods) on NTU-60 dataset X-Sub, with different combinations of frame-masking ($\alpha$) and joint-masking ($\beta$) ratios. We set the coordinates ($\alpha$, $\beta$) as the values of x-axis, with a descending order of the $\alpha$. We observe that the overall results of the random masking are better than the fixed index masking. Specifically, the random masking method surpasses the fixed index masking method with the largest gap ($1.6\%$) at (0.5,0.5), and there exists a smallest gap ($0.2\%$) at (0.4,0.6).
%%%%%%%%%%%%%%%%%%%%%%%%%%%%%%%%%%%%%%%%%%%%%%%%%%%%%%%%%%%%
%%%%%%%%%%%%%%%%%%%%%%%%%%%%%%%%%%%

\clearpage

\begin{table}[h]
\vspace{-5pt}
  \resizebox{1\linewidth}{!}
{
    \begin{tabular}{|l|l|l|l|l|}
    \hline
        SkeletonMAE & layer name & input dim ($D_{in}$) & output dim ($D_{out}$) & QKV dim \\ \hline
        ~ & input layer & 3 & 64 & ~ \\ 
        ~ & Block1 & 64 & 64 & 16 \\ 
        ~ & Block2 & 64 & 64 & 16 \\ 
        ~ & Block3 & 64 & 128 & 32 \\ 
        encoder & Block4 & 128 & 128 & 32 \\ 
        ~ & Block5 & 128 & 256 & 64 \\ 
        ~ & Block6 & 256 & 256 & 64 \\ 
        ~ & Block7 & 256 & 256 & 64 \\ 
        ~ & Block8 & 256 & 256 & 64 \\ \hline
        ~ & Block1 & 256 & 256 & 64 \\ 
        ~ & Block2 & 256 & 256 & 64 \\ 
        ~ & Block3 & 256 & 256 & 64 \\ 
        ~ & Block4 & 256 & 128 & 64 \\ 
decoder & Block5 & 128 & 128 & 32 \\ 
        ~ & Block6 & 128 & 64 & 32 \\ 
        ~ & Block7 & 64 & 64 & 16 \\ 
        ~ & Block8 & 64 & 64 & 16 \\ 
        ~ & output layer & 64 & 3 & 16 \\ \hline
    \end{tabular}
}
\caption{The detailed structures of SkeletonMAE encoder and decoder, where the embedding dimension is 256 and the depth of decoder is 9.}
\label{tab: table1}
\vspace{-5pt}
\end{table}
%%%%%%%%%%%%%%%%%%%%%%%%%%%%%%%%%%%%%%%%%%%%%%%%%%%%%%%%%%%%
\begin{table}[h]
\vspace{-5pt}
  \resizebox{1\linewidth}{!}
{
    \begin{tabular}{|l|l|l|l|l|}
    \hline
        SkeletonMAE & layer name & input dim ($D_{in}$) & output dim ($D_{out}$) & QKV dim \\ \hline
        ~ & input layer & 3 & 64 & ~ \\ 
        ~ & Block1 & 64 & 64 & 16 \\ 
        ~ & Block2 & 64 & 128 & 32 \\ 
        ~ & Block3 & 128 & 128 & 32 \\ 
encoder & Block4 & 128 & 256 & 64 \\ 
        ~ & Block5 & 256 & 256 & 64 \\ 
        ~ & Block6 & 256 & 512 & 128 \\ 
        ~ & Block7 & 512 & 512 & 128 \\ 
        ~ & Block8 & 512 & 512 & 128 \\ \hline
        ~ & Block1 & 512 & 512 & 128 \\ 
        ~ & Block2 & 512 & 512 & 128 \\ 
        ~ & Block3 & 512 &256 & 64 \\ 
        ~ & Block4 & 256 & 256 & 64 \\ 
decoder & Block5 & 256 & 128 & 32 \\ 
        ~ & Block6 & 128 & 128 & 32 \\ 
        ~ & Block7 & 128 & 64 & 16 \\ 
        ~ & Block8 & 64 & 64 & 16 \\ 
        ~ & output layer & 64 & 3 & 16 \\ \hline
    \end{tabular}
}
\caption{The detailed structures of SkeletonMAE encoder and decoder, where the embedding dimension is 512 and the depth of decoder is 9.}
\label{tab: table2}
\vspace{-5pt}
\end{table}

%%%%%%%%%%%%%%%%%%%%%%%%%%%%%%%%%%%%%%%%%%%%%%%%%%%%%%%%%%%%
\begin{table}[h]
\vspace{-5pt}
  \resizebox{1\linewidth}{!}
{
\begin{tabular}{|l|l|l|l|l|}
    \hline
        SkeletonMAE & layer name & input dim ($D_{in}$) & output dim ($D_{out}$) & QKV dim \\ \hline
        ~ & input layer & 3 & 32 & ~ \\ 
        ~ & Block1 & 32 & 32 & 8 \\ 
        ~ & Block2 & 32 & 32 & 8 \\ 
        ~ & Block3 & 32 & 64 & 16 \\ 
encoder & Block4 & 64 & 64 & 16 \\ 
        ~ & Block5 & 64 & 128 & 32 \\ 
        ~ & Block6 &128 & 128 & 32 \\ 
        ~ & Block7 &128 & 128 & 32 \\ 
        ~ & Block8 &128 & 128 & 32 \\ \hline
        ~ & Block1 &128 & 128 & 32 \\ 
        ~ & Block2 &128 & 128 & 32 \\ 
        ~ & Block3 &128 & 128 & 32 \\ 
        ~ & Block4 &128 & 64 & 32 \\ 
decoder & Block5 & 64 & 64 & 16 \\ 
        ~ & Block6 & 64 & 32 & 16 \\ 
        ~ & Block7 & 32 & 32 & 8 \\ 
        ~ & Block8 & 32 & 32 & 8 \\ 
        ~ & output layer & 32 & 3 & 8 \\ \hline
    \end{tabular}
}
\caption{The detailed structures of SkeletonMAE encoder and decoder, where the embedding dimension is 128 and the depth of decoder is 9.}
\label{tab: table3}
\vspace{-5pt}
\end{table}

%%%%%%%%%%%%%%%%%%%%%%%%%%%%%%%%%%%%%%%%%%%%%%%%%%%%%%%%%%%%
\begin{table}[h]
\vspace{-5pt}
  \resizebox{1\linewidth}{!}
{
\begin{tabular}{|l|l|l|l|l|}
    \hline
        SkeletonMAE & layer name & input dim ($D_{in}$) & output dim ($D_{out}$) & QKV dim \\ \hline
        ~ & input layer & 3 & 64 & ~ \\ 
        ~ & Block1 & 64 & 64 & 16 \\ 
        ~ & Block2 & 64 & 64 & 16 \\ 
        ~ & Block3 & 64 & 128 & 32 \\ 
encoder & Block4 & 128 & 128 & 32 \\ 
        ~ & Block5 & 128 & 256 & 64 \\ 
        ~ & Block6 & 256 & 256 & 64 \\ 
        ~ & Block7 & 256 & 256 & 64 \\ 
        ~ & Block8 & 256 & 256 & 64 \\ \hline
        ~ & Block1 & 256 & 128 & 64 \\ 
~& Block2 & 128 & 128 & 32 \\ 
       decoder & Block3 & 128 & 64 & 32 \\ 
        ~ & Block4 & 64 & 64 & 16 \\ 
        ~ & output layer & 64 & 3 & 16 \\ \hline
    \end{tabular}
}
\caption{The detailed structures of SkeletonMAE encoder and decoder, where the embedding dimension is 256 and the depth of decoder is 5.}
\label{tab: table4}
\vspace{-5pt}
\end{table}

%%%%%%%%%%%%%%%%%%%%%%%%%%%%%%%%%%%%%%%%%%%%%%%%%%%%%%%%%%%%
\begin{table}[h]
\vspace{-5pt}
  \resizebox{1\linewidth}{!}
{
\begin{tabular}{|l|l|l|l|l|}
    \hline
        SkeletonMAE & layer name & input dim ($D_{in}$) & output dim ($D_{out}$) & QKV dim \\ \hline
        ~ & input layer & 3 & 64 & ~ \\ 
        ~ & Block1 & 64 & 64 & 16 \\ 
        ~ & Block2 & 64 & 64 & 16 \\ 
        ~ & Block3 & 64 & 128 & 32 \\ 
encoder & Block4 & 128 & 128 & 32 \\ 
        ~ & Block5 & 128 & 256 & 64 \\ 
        ~ & Block6 & 256 & 256 & 64 \\ 
        ~ & Block7 & 256 & 256 & 64 \\ 
        ~ & Block8 & 256 & 256 & 64 \\ \hline
        ~ & Block1 & 256 & 256 & 64 \\ 
        ~ & Block2 & 256 & 256 & 64 \\ 
        ~ & Block3 & 256 & 128 & 64 \\ 
decoder & Block4 & 128 & 128 & 32 \\ 
        ~ & Block5 & 128 & 64 & 32 \\ 
        ~ & Block6 & 64 & 64 & 16 \\ 
        ~ & output layer & 64 & 3 & 16 \\ \hline
    \end{tabular}
}
\caption{The detailed structures of SkeletonMAE encoder and decoder, where the embedding dimension is 256 and the depth of decoder is 7.}
\label{tab: table5}
\vspace{-5pt}
\end{table}

%%%%%%%%%%%%%%%%%%%%%%%%%%%%%%%%%%%%%%%%%%%%%%%%%%%%%%%%%%%%
\begin{table}[h]
\vspace{-5pt}
  \resizebox{1\linewidth}{!}
{
\begin{tabular}{|l|l|l|l|l|}
    \hline
        SkeletonMAE & layer name & input dim ($D_{in}$) & output dim ($D_{out}$) & QKV dim \\ \hline
        ~ & input layer & 3 & 64 & ~ \\ 
        ~ & Block1 & 64 & 64 & 16 \\ 
        ~ & Block2 & 64 & 64 & 16 \\ 
        ~ & Block3 & 64 & 128 & 32 \\ 
encoder & Block4 & 128 & 128 & 32 \\ 
        ~ & Block5 & 128 & 256 & 64 \\ 
        ~ & Block6 & 256 & 256 & 64 \\ 
        ~ & Block7 & 256 & 256 & 64 \\ 
        ~ & Block8 & 256 & 256 & 64 \\ \hline
        ~ & Block1 & 256 & 256 & 64 \\ 
        ~ & Block2 & 256 & 256 & 64 \\ 
~ & Block3 & 256 & 256 & 64 \\
        ~ & Block4 & 256 & 256 & 64 \\ 
        ~ & Block5 & 256 & 128 & 64 \\ 
decoder & Block6 & 128 & 128 & 32 \\ 
~& Block7 & 128 & 128 & 32 \\
        ~ & Block8 & 128 & 64 & 32 \\ 
        ~ & Block9 & 64 & 64 & 16 \\ 
        ~ & Block10 & 64 & 64 & 16 \\ 
        ~ & output layer & 64 & 3 & 16 \\ \hline
    \end{tabular}
}
\caption{The detailed structures of SkeletonMAE encoder and decoder, where the embedding dimension is 256 and the depth of decoder is 11.}
\label{tab: table6}
\vspace{-5pt}
\end{table}

%%%%%%%%%%%%%%%%%%%%%%%%%%%%%%%%%%%%%%%%%%%%%%%%%%%%%%%%%%%%

%%%%%%%%%%%%%%%%%%%%%%%%%%%%%%%%%%%%%%%%%%%%%%%%

\end{document}